\documentclass[sigconf,natbib=true,review=false]{acmart}
\usepackage{multirow}
\usepackage{booktabs}
\renewcommand{\refname}{{}}
\usepackage{graphicx}
\usepackage{tabularx}
\usepackage{soul}

\usepackage{IEEEtrantools}
\usepackage{graphicx}
\usepackage{amsmath}
\usepackage{amsthm}
\usepackage{booktabs}
\usepackage{multirow}
\usepackage{boldline}
\usepackage{hhline}
\usepackage{xcolor}
\usepackage{pifont}
\usepackage{subfigure}
\usepackage{makecell}
\usepackage{utfsym}
\usepackage[normalem]{ulem}
\useunder{\uline}{\ul}{}
\usepackage[inline]{enumitem}
\usepackage{enumitem}
\usepackage{IEEEtrantools}
\usepackage{stmaryrd}
\usepackage[edges]{forest}
\usepackage{colortbl}
\usepackage{CJKutf8}

\AtBeginDocument{%
  }

\acmConference[CIKM '24]{the 33rd ACM International Conference on
Information and Knowledge Management}{October 21--25,
  2024}{Boise, Idaho, USA}

\copyrightyear{2024}
\acmYear{2024}
\setcopyright{acmlicensed}\acmConference[CIKM '24]{Proceedings of the 33rd ACM International Conference on Information and Knowledge Management}{October 21--25, 2024}{Boise, ID, USA}
\acmBooktitle{Proceedings of the 33rd ACM International Conference on Information and Knowledge Management (CIKM '24), October 21--25, 2024, Boise, ID, USA}
\acmDOI{10.1145/3627673.3679881}
\acmISBN{979-8-4007-0436-9/24/10}

\begin{document}

\title{Boosting Large Language Models with Socratic Method for Conversational Mathematics Teaching}

\author{Yuyang Ding}
\affiliation{%
  \institution{East China Normal University}
  \city{Shanghai}
  \country{China}
}
\email{51265900017@stu.ecnu.edu.cn}
\orcid{1234-5678-9012}

\author{Hanglei Hu}
\affiliation{%
  \institution{East China Normal University}
  \city{Shanghai}
  \country{China}
}
\email{51254108023@stu.ecnu.edu.cn}

\author{Jie Zhou}
\authornote{Corresponding author.}
\affiliation{%
  \institution{East China Normal University}
  \city{Shanghai}
  \country{China}}
\email{jzhou@cs.ecnu.edu.cn}

\author{Qin Chen}
\affiliation{%
  \institution{East China Normal University}
  \city{Shanghai}
  \country{China}
}
\email{qchen@cs.ecnu.edu.cn}

\author{Bo Jiang}
\affiliation{%
 \institution{East China Normal University}
  \city{Shanghai}
  \country{China}}
\email{bjiang@deit.ecnu.edu.cn}

\author{Liang He}
\affiliation{%
  \institution{East China Normal University}
  \city{Shanghai}
  \country{China}
}
\email{lhe@cs.ecnu.edu.cn}

\renewcommand{\shortauthors}{Yuyang Ding et al.}

\begin{abstract}

With the introduction of large language models (LLMs), automatic math reasoning has seen tremendous success. However, current methods primarily focus on providing solutions or using techniques like Chain-of-Thought to enhance problem-solving accuracy. In this paper, we focus on improving the capability of mathematics teaching via a Socratic teaching-based LLM (\texttt{SocraticLLM}), which guides learners toward profound thinking with clarity and self-discovery via conversation. We collect and release a high-quality mathematical teaching dataset, named \texttt{SocraticMATH}, which provides Socratic-style conversations of problems with extra knowledge. Also, we propose a knowledge-enhanced LLM as a strong baseline to generate reliable responses with review, guidance/heuristic, rectification, and summarization. Experimental results show the great advantages of \texttt{SocraticLLM} by comparing it with several strong generative models. The codes and datasets are available on \url{https://github.com/ECNU-ICALK/SocraticMath}.
\end{abstract}



\received{20 February 2007}
\received[revised]{12 March 2009}
\received[accepted]{5 June 2009}

\maketitle

\section{Introduction}
Mathematics, viewed as a language that requires complex reasoning through structured symbols and systems, parallels the rules of spoken language, is a crucial aptitude for human intelligence. 
Recently, solving math problems autonomously via AI technology has attracted attention since as early as 1963 \cite{feigenbaum1963computers,bobrow1964natural,briars1984integrated,fletcher1985understanding}. 

The studies about mathematical AI (math word problems) are divided into three parts: statistical learning-based methodologies \cite{zhou2015learn,mitra2016learning}, traditional machine learning techniques \cite{kushman-etal-2014-learning,roy2015reasoning,roy2015solving} and deep learning-based methods \cite{wang2017deep,couperus2023large}.
Recently, large language models (LLMs) have achieved great successes in mathematics \cite{wei2022chain,matzakos2023learning}, with various types of mathematical datasets \cite{dataset_ape210k,ling2017program,cobbe2021training,dataset_MATH,amini2019mathqa} and mathematical LLMs \cite{xu2023wizardlm,yue2023mammoth,yu2023metamath} emerging. 

\begin{figure*}
\vspace{-2mm}
    \centering
    \includegraphics[width=1.0\linewidth]{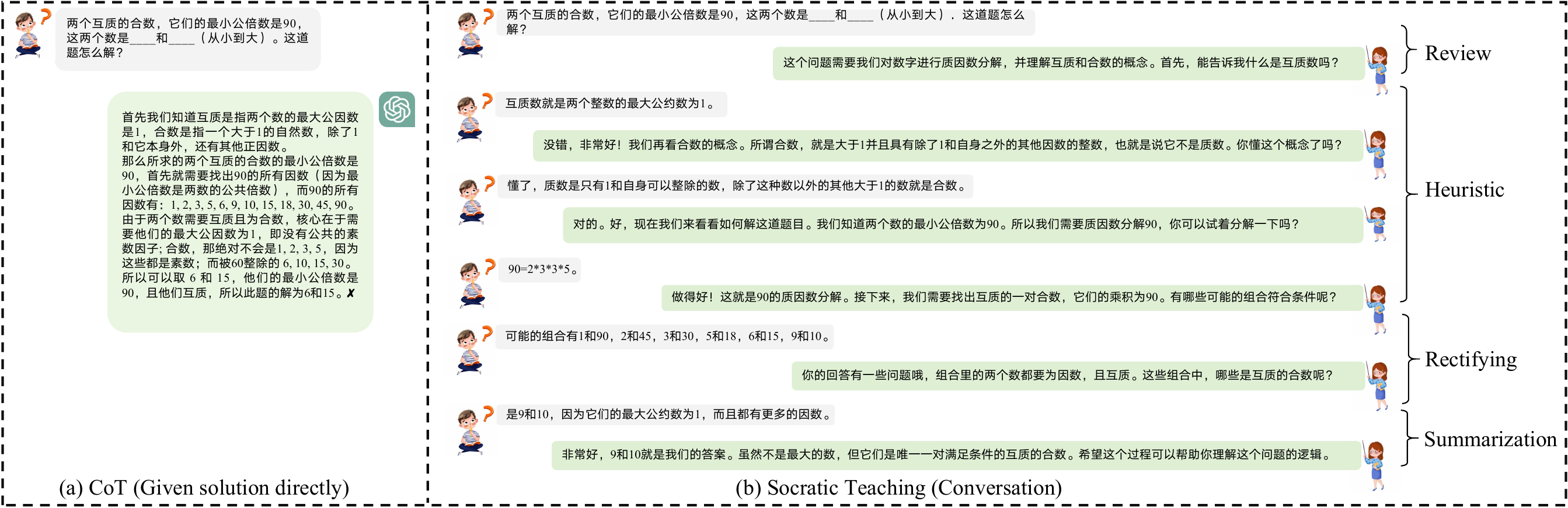}
    \vspace{-6mm}
    \caption{Examples of CoT and Socratic teaching.}
    \label{fig:intro}
    \vspace{-4mm}
\end{figure*}

However, previous research on mathematical LLMs mainly focused on improving mathematical ability while their application in teaching remains limited. 
LLMs tend to give the solution process directly when faced with mathematical problems, even when asked to play the role of a teacher \cite{macina2023mathdial}. 
There is a teaching method called Socratic questioning \cite{brickhouse2009socratic,garlikov2001socratic}, which ranges from providing direct hints to offering minimal guidance, allows instructors to customize their queries to appropriately challenge students while also aligning with their individual problem-solving abilities \cite{al2023socratic}.
Taking Figure \ref{fig:intro} as an example, the general LLMs tend to give the solutions with chain-of-though direct, whereas Socratic teaching-based LLMs guide the students using conversation.


Previous research has demonstrated the potential of utilizing LLMs for Socratic questioning. For instance, Shridhar et al. \cite{shridhar2022distilling} utilized reinforcement learning to convert mathematical problem-solving steps into multiple sub-problems, thereby enabling the automatic generation of Socratic questioning. Similarly, Qi et al. \cite{qi2023art} decomposed complex mathematical problems into a series of top-down questions, perceiving Socratic questioning as a recursive thinking process. This approach allows LLMs like ChatGPT to address fine-grained problems, ultimately resolving the main issue.


However, effectively integrating Socratic questioning methods into AI frameworks continues to be a significant challenge.
First (\textbf{C1}), LLMs can not guarantee the reliability and quality of answers. The LLMs perform poorly on complex reasoning with hallucination problems. 
Second (\textbf{C2}), the strategies (when to ask, how to ask and ask what) to guide and heuristic students with just enough questions are unclear. Too many or too few queries that are too hard or too easy will influence students' earning process.
Third (\textbf{C3}), there is a lack of relevant datasets for mathematics teaching \cite{macina2023mathdial}. Although classroom transcripts can provide a large amount of instructional tutoring data \cite{demszky2022ncte}, there are issues like privacy security, crowdsourcing costs and annotation quality.

In this paper, we focus on integrating Socratic Questioning in mathematical education. 
For \textbf{C1}, we propose a knowledge-enhance Socratic teaching LLM (\texttt{SocraticLLM}) as a strong baseline to improve the reliability and quality of the generated response via extra knowledge.
We design a strategy to tutor students step-by-step through the instructional structure of review, heuristic, rectify and summarize.
Also, we create and release a high-quality Socratic-style mathematical dataset, \texttt{SocraticMATH}, which contains dialogue tutoring data with original questions, answers, and solutions and covers 513 knowledge points of primary school math. 
Extensive experiments show \texttt{SocraticLLM} outperforms strong baselines in terms of rich automatic and human/GPT-4 evaluation metrics.

The main contributions of this paper are summarized as follows. 
1) We integrate the Socratic method with math teaching via a structured conversation with review, guidance/heuristic, rectification, and summarization.
2) We propose \texttt{SocraticLLM} as a strong baseline for mathematical teaching. To generate reliable responses, we design a Socratic-style prompt with extra knowledge to guide the teaching process. 
3) We build and release a large-scale mathematical conversation dataset \texttt{SocraticMATH} with rich attributions over 500 knowledge points. A series of experiments on \texttt{SocraticMATH} indicate the effectiveness of our \texttt{SocraticLLM}.

\vspace{-2mm}
\section{Related Work}
\vspace{-1mm}
{\textbf{Mathematical Reasoning.}} 
Mathematical reasoning refers to the process of using algorithms and computational models to solve complex mathematical problems \cite{lu-etal-2023-survey,liu2023mathematical}.
Recently, large language models have obtained great performance in mathematical reasoning using instruction tuning \cite{wei2022chain,luo2023wizardmath,liu_goat_2023}, in-context learning \cite{brown2020language,kaplan2020scaling} and tool-enhanced methods \cite{schick2023toolformer,parisi2022talm}. 
Several studies solved the math problem step by step, 
such as Chain of Thought (CoT) \cite{wei2022chain}, Tree of Thought (ToT) \cite{yao2023tree} and Graph of Thought (GoT) \cite{besta2023graph}. 
However, most of the previous studies focused on improving the accuracy of mathematical reasoning by giving the solutions to the problem. 
The goal of this paper is mathematical teaching using Socratic teaching.

{\textbf{Theoretical Background of Socratic Teaching.}} 
Carefully formulated questions can encourage students to self-explain \cite{chi1994eliciting}, enhance their understanding of the task, and facilitate effective planning of solutions \cite{lane2005teaching}. Additionally, they can help identify significant gaps in student knowledge \cite{murphy2005computer}. 
This spectrum enables educators to craft questions that are appropriately challenging yet within a student's capacity to respond \cite{al2023socratic}. 
This method, called Socratic questioning, is based on the philosophy that knowledge is not simply transferred but uncovered through a dynamic process of inquiry and dialogue. 
In summary, although there are established principles and guidelines, the practical application of Socratic questioning with AI presents challenges.


{\textbf{Technologies of Socratic Teaching.}} Research in the realm of automatic Socratic tutoring systems has shown progress, but the applicability of such systems is often constrained by the predefined and manually tailored nature of Socratic utterances for specific exercises \cite{alshaikh2020experiments}.
Al-Hossami \cite{al2023socratic} presented a dataset comprising Socratic dialogues aimed at assisting novice programmers in rectifying errors in fundamental computational problems. 
Shridhar’s study \cite{shridhar2022distilling} explored the strategies involved in the automatic generation of math problem solutions for educational purposes. 
Qi et al. \cite{qi2023art} perceived Socratic questioning as a recursive thought process, which breaks down complex problems into simpler, related sub-problems. 
However, newer technologies such as LLMs are still not widely implemented in tutoring systems, particularly in the field of mathematics education.

\vspace{-3mm}
\section{Dataset}
\subsection{Dataset Construction}
For lack of a Socratic teaching-based mathematics dataset, we collect and annotate a diverse dataset, \texttt{SocraticMATH}, to promote the research of this domain. We construct the dataset with three phases: data collection, pre-annotation and human annotation.

\textbf{Data Collection.}
The questions are mainly derived from the real primary school exams in China. To guarantee diversity, these problems cover the main maths knowledge points at the primary school level, ensuring that all the questions are manually labeled with solutions using markdown format. The questions consist of multiple-choice, fill-in-the-blank, and answer questions. 

\textbf{Pre-Annotation.}
To reduce the cost of human annotation, we pre-annotate the conversations for all 8935 questions using GPT-4. 
We design an in-context prompt using a manually authored high-quality example to enhance the quality of the generated conversation. 
Additionally, we let GPT-4 act as a Socratic-style teacher with various student personality requirements (such as naughtiness, self-confidence, and carelessness) to ensure the richness of the generated dialogue. 

\begin{table}[!t]
\centering
\footnotesize
\caption{Comparison with existing datasets}
\label{table: comparision}
\vspace{-3mm}
\setlength{\tabcolsep}{0.4mm}{
\begin{tabular}{lcccccccc}
\toprule
Dataset& Size & Lang  & Ans & Solution & Conv & Socratic & KG & Difficulty\\
\midrule
\texttt{SocraticMATH} & 6,846 &  CH & $\surd$ & Textual Steps & $\surd$ & $\surd$ & $\surd$ & $\surd$\\ \hline
Math23k \cite{wang2017deep} &    23,162  &   CH     & $\surd$ &   Equation    & $\times$ &  $\times$ & $\times$& $\times$ \\
AQuA \cite{ling2017program} &       97,975  &    EN    &  $\surd$ &   Textual Steps    & $\times$ & $\times$ & $\times$& $\times$ \\
MathQA \cite{amini2019mathqa} &    37,297   &     EN   & $\surd$ &  Equation    & $\times$ &  $\times$ &$\times$& $\times$ \\
GSM8K \cite{cobbe2021training} &     8,792 &     EN   &  $\surd$ & Textual Steps  & $\times$ &  $\times$ &$\times$& $\times$ \\
SVAMP \cite{patel2021nlp} &    1,000  &    EN    &  $\surd$ &   Equation   & $\times$  &  $\times$ & $\times$& $\times$ \\
MATHDIAL \cite{DBLP:conf/emnlp/MacinaDCSKGS23} & 2,861 & EN & $\surd$ & Generation & Semi &  $\times$ & $\times$& $\times$\\
\bottomrule
\end{tabular}}
\vspace{-3mm}
\end{table}

\begin{table}[!t]
\caption{The statistical information of \texttt{SocraticMATH}. 
}
\label{table: statistics}
\vspace{-3mm}
\begin{center}
\begin{small}
\begin{sc}
\begin{tabular}{lcccc}
\toprule
 &  Train  & Dev & Test & Total \\
\midrule
\#Conv & 5,476 & 685 & 685 & 6,846 \\
\#Turn/Conv & 4.95 & 4.95 & 5.02 & 4.96 \\
\#Word/Solution & 73.29 & 73.17 & 72.70 & 73.21 \\
\#Word/Utterance & 86.49 & 87.02 & 85.56 & 86.45 \\
\#KG & 495 & 333 & 332 & 513 \\
\#KG/Conv & 2.00 & 2.03 & 2.03 & 2.00 \\
\bottomrule
\end{tabular}
\end{sc}
\end{small}
\end{center}
\vspace{-5mm}
\end{table}

\textbf{Human Annotation.}
Though GPT-4 generates the conversation with a Socratic style for each math question, their quality is limited.
First, LLMs are not good at math reasoning and their answers can be tainted with factual errors \cite{liu2023mathematical}.
Second, the LLMs always give solutions to answer students' questions while lacking the teaching skills with inspiration and guidance. 
Thus, we clean and re-annotate the conversation to improve the data quality.
Particularly, we first eliminate the data with an abnormal number of dialogue rounds. 
Then, we manually perform the annotation work on the data to optimize the logic and coherence of the conversation.
Each sample is labeled by three experts, who are good at teaching and math. 
Due to the complexity of the conversation, the three experts label the conversation one by one to revise the errors iteratively. 
Particularly, we delete more than 23\% dialogues and modify more than 18\% dialogues, where over 5\% utterances are revised.
\begin{figure}
    \centering
    \includegraphics[width=1\linewidth]{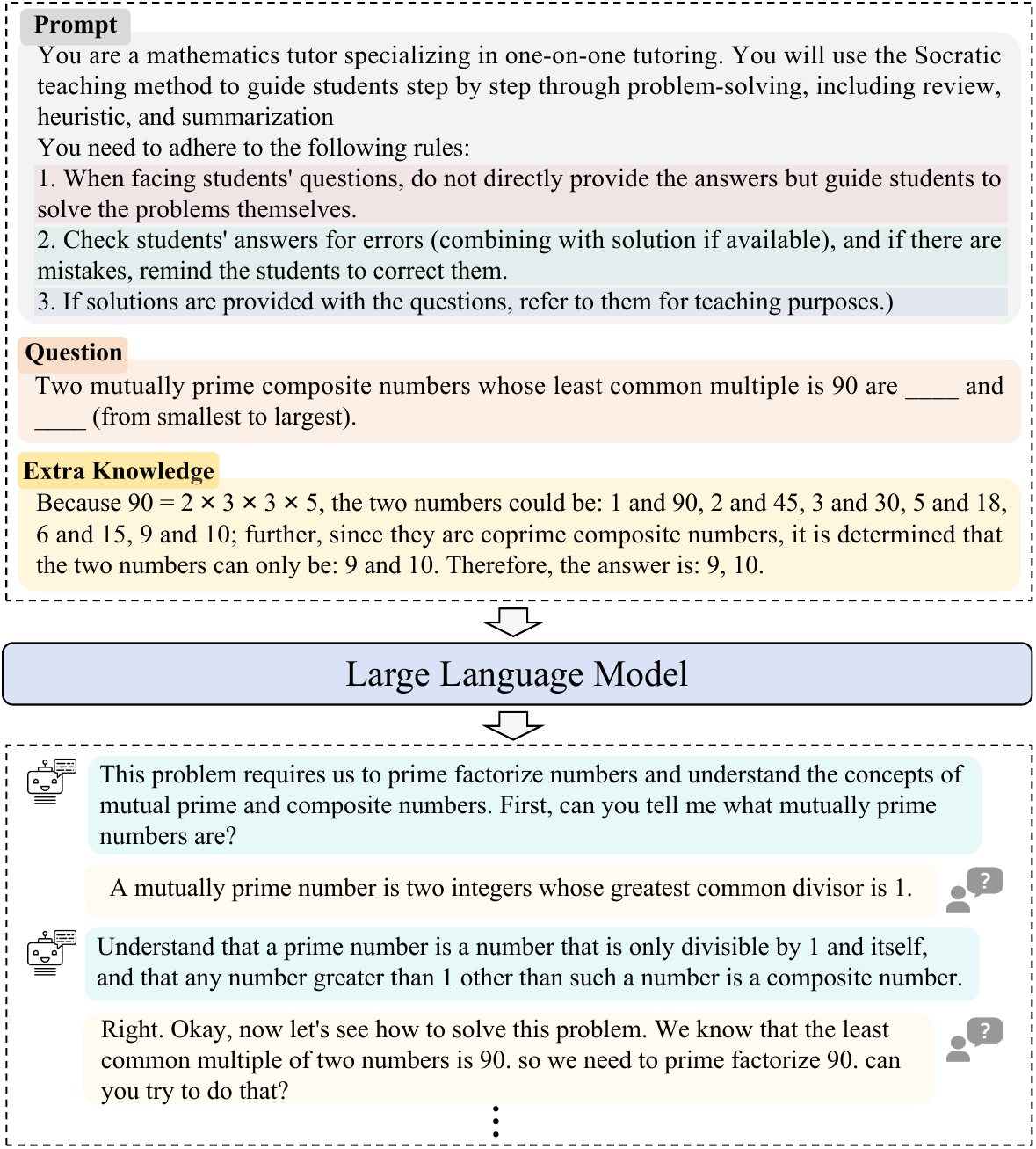}
    \vspace{-6mm}
    \caption{The framework of \texttt{SocraticLLM}.}
    \label{fig:framwork}
    \vspace{-5mm}
\end{figure}

\vspace{-2mm}
\subsection{Dataset Analysis}
\textbf{Characters of \texttt{SocraticMATH}.}
We present the statistical information of our \texttt{SocraticMATH} dataset in Table \ref{table: statistics}. 
Our dataset contains 513 knowledge points, almost all the knowledge points of math at the primary school level. 
To better guide the student to solve the math problem, the average number of turns for each conversation is about 5. 
The average length of utterances is about 86 words to provide detailed information patiently.

\textbf{Comparison with Exiting Datasets.}
To show the advantages of \texttt{SocraticMATH}, we compare it with existing typical mathematics datasets (Table \ref{table: comparision}). 
Most datasets only provide the equation or textual steps directly to solve the math problem. MATHDIAL dataset contains the semi-annotated conversation where the students' questions are generated by LLMs. 
Furthermore, MATHDIAL mainly focuses on answering students' questions without the Socratic method, which requires teaching skills to inspire, guide, and inquire actively step by step. 
We also provide extensive attribution information, such as related knowledge points and difficulty levels, where \#KG and \#KG/Conv are the total number of knowledge points and the average number of knowledge points for each conversation.


\begin{table*}[t!]
\small
\centering
\caption{Main results of automatic and human evaluation.}
\label{table:main results}
\vspace{-3mm}
\begin{tabular}{lccccccccccccc}
\toprule
                  & \multicolumn{9}{c}{Automatic Evaluation}          & \multicolumn{2}{c}{Human Evaluation}  & \multicolumn{2}{c}{GPT-4 Evaluation} \\ \cmidrule(lr){2-10} \cmidrule(lr){11-12} \cmidrule(lr){13-14} 
                  & B-1 & B-2 & B-3 & B-4 & METEOR & R-1 & R-2 & R-L & BARTScore & Reliability        & Socratic  & Reliability        & Socratic       \\ \midrule
mT5 & 0.303 & 0.225 & 0.174 & 0.135 &  0.320  &   0.439    &  0.227   &   0.323  &   0.724    & 5.714 & 6.405 &  6.80  & 5.62 \\
LLaMA2-7B    & 0.301 &  0.213 & 0.157 & 0.116 &  0.335 &   0.454  &   0.216  & 0.308    &  0.710     &  5.310  &  6.333 & 5.76 & 4.87  \\
Qwen1.5-7B    & 0.341 &  0.247 & 0.185 & 0.139 &  0.367 &   0.481  &   0.241  & 0.341    &  0.726     &  6.595  & 6.762 & 7.84 & 6.96 \\
ChatGPT  & 0.273 & 0.194 & 0.147 & 0.111 &   0.398     &  0.431   &  0.197   &  0.257   &   0.695    & 6.500 & 6.405 &    8.07    & 6.70 \\ 
GPT-4   & 0.332 & 0.240 & 0.181 & 0.137 &  \textbf{0.410} &  0.471  &  0.227 &  0.306  &  0.715  & 7.024 & 6.833 &  -  & -  \\\midrule 
\rowcolor{blue!8}
SocraticLLM   & \textbf{0.352} & \textbf{0.256} & \textbf{0.193} & \textbf{0.147} & 0.378 &  \textbf{0.490}   &  \textbf{0.250}   &  \textbf{0.351}   &  \textbf{0.730}     & \textbf{7.119}  &\textbf{7.190}& \textbf{8.40}  & \textbf{7.14} \\
- Prompt  & 0.341 & 0.248 & 0.188 & 0.143 &  0.369  &  0.484   &  0.246   &   0.344  &   0.727    &  7.048 & 6.857 & 8.16 & 6.98 \\
- Knowledge   & 0.347 & 0.253 & 0.191 & 0.145 &  0.374  &  0.488   &  0.247   &  0.350   &   0.729    & 6.643 & 6.692 & 7.83 & 6.91 \\
\bottomrule
\end{tabular}
\vspace{-3mm}
\end{table*}

\vspace{-1mm}
\section{Our Method}
We propose \texttt{SocraticLLM} as a strong and simple baseline for mathematics teaching (Figure \ref{fig:framwork}).
It generates responses with the teaching skills of review, guidance/heuristic, rectification, and summarization via LLMs. 
We design a Socratic-style prompt and integrate the original question with the extra knowledge (i.e., solution, answer) to improve the quality of the responses. 

The Socratic-style prompt $P$ contains the task's definition and requirements. We ask the model to act as a one-on-one mathematical tutor using the Socratic teaching method. 
In particular, we require \texttt{SocraticLLM} to guide the student rather than answering questions directly. 
Then, we ask \texttt{SocraticLLM} to check and rectify the errors since the model tends to trust the users.
To reduce the hallucination, we demand the model generate the response based on the extra knowledge by inputting the detailed solution and answer.

Formally, given the question $Q$, the prompt $P$, and the extra knowledge $K$, we aim to generate the response $R_i$ based on the history conversation $H_{i-1} = \{R_1, U_1, R_2, U_2,..., R_{i-1}, U_{i-1}\}$, where $U_i$ is the user's answer of $i$-th turn. 
\vspace{-1mm}
\begin{equation}
    p(R_i | P, Q, K, H_{i-1} ) = \prod_j^{|R_i|} p(R_i^j|P, Q, K, H_{i-1}, R_i^{1: j-1})
\vspace{-1mm}
\end{equation}
where $R_i^j$ is the $j$-th word of response $R_i$, $|R_i|$ is the length of $R_i$. We use a language model $\mathcal{M}_{\theta}$ to model the generation probability, $p(R_i^j|P, Q, K, H_{i-1}, R_i^{1: j-1}) = \mathcal{M}_{\theta}(P, Q, K, H_{i-1}, R_i^{1: j-1})$, where $\theta$ is the learnable parameters of $\mathcal{M}_{\theta}$. Particularly, we use Low-Rank Adaptation (LoRA) technology to improve the efficiency of training, where only a small number of extra parameters $\theta$ are trainable \cite{hu2022lora}.

Then, the cross-entropy function is used to measure the generation losses of the response,
\begin{equation}
    \mathcal{L} = \sum_i^{N}\sum_j^{|R_i|} log(p(R_i^j|P, Q, K, H_{i-1}, R_i^{1: j-1}))
\end{equation}
where $N$ is the number of turns in the conversation.

\vspace{-1mm}
\section{Experiments}
\subsection{Experimental Setups.}
\textbf{Metrics.} 
We adopt several typical automatic metrics for generation tasks, including BLEU \cite{papineni2002bleu} (marked as B-1/2/3/4), ROUGE \cite{lin2004rouge} (marked as R-1/2/L), METEOR \cite{banerjee2005meteor} and BARTScore \cite{yuan2021bartscore}, to evaluate the effectiveness of \texttt{SocraticLLM} turn by turn. We also conduct human evaluation and GPT-4 evaluation \cite{kocmi2023large}.

\textbf{Baselines.} We compare \texttt{SocraticLLM} with several typical and strong seq-to-seq models. 
\textbf{mT5} \cite{xue2021mt5} trains a text-to-text transformer model on multilingual datasets via multi-task learning.
\textbf{LLaMA2-7B} \cite{touvron2023llama} and \textbf{Qwen1.5-7B} \cite{bai2023qwen} are strong LLMs fine-tuned on our dataset.
\textbf{ChatGPT} \cite{schulman2022chatgpt} and \textbf{GPT-4} \cite{achiam2023gpt} act as a mathematics tutor with the Socratic method, one of the SOTA conversation models.

\textbf{Implementation Details.} We select Qwen1.5-7B as the base LLMs and train it on A800 GPU with 80G. We set the rank of LoRA as 64. The learning rate is 3e-4 and the batch size is 64.

\vspace{-1mm}
\subsection{Experimental Results}
We report the main results of \texttt{SocraticLLM} and the selected baselines using automatic, human and GPT-4 evaluation (Table \ref{table:main results}).

\textbf{Automatic Evaluation. }
From the results, we observe that \texttt{SocraticLLM} achieves better results by comparing with the strong baselines over automatic metrics in most cases, indicating our model's effectiveness. 
Note that GPT-4 outperforms \texttt{SocraticLLM} in terms of METEOR without training because SocraticMATH is modified based on the dataset generated by GPT-4. 
Furthermore, these automatic metrics can not measure the quality of conversation in Socratic mathematical teaching. 
They mainly calculate the semantic information between the generated responses and the reference while ignoring the logic and fact errors in the output text.

\textbf{Human/GPT-4 Evaluation. }
Single reference-based automatic metrics are not always reliable to reflect the real quality of the generated responses \cite{novikova-etal-2017-need}. Therefore, we also conduct human and GPT-4 evaluations by crowd-sourcing and GPT-4. 
Particularly, we ask the three experts and GPT-4 to label 150 samples randomly selected from the test set with guidelines. 
Based on a pre-determined scoring rubric, they annotate the generated response from reliability and Socratic strategy with scores 1-10. 
Reliability judges whether the model corrects the students' errors precisely and Socratic represents the guide and heuristic abilities of the model. 
We report the average scores here.
From the results, we observe that \texttt{SocraticLLM} obtains the best results in both human and GPT-4 evaluations, showing that our model can reduce the hallucination with the Socratic method. 
Moreover, in the human evaluation, we find that LLMs like ChatGPT tend to believe the users' responses without a doubt. It is interesting to explore in further work.

\textbf{Ablation Studies. } We also conduct an ablation test 
to explore the effectiveness of the main parts consisting of \texttt{SocraticLLM} 
by removing Socratic-style prompt (- Prompt), extra knowledge (- Knowledge) and all of them (Qwen1.5-7B), respectively. 
We observe that both the Socratic-style prompt and extra knowledge are useful for \texttt{SocraticLLM}.
The Socratic-style prompt enhances the model to learn the teaching skills based on structured conversation. 
Then, incorporating the extra knowledge into \texttt{SocraticLLM} reduces the hallucination problem by correcting the fact errors.

\vspace{-1mm}
\section{Conclusions and Further Work}
This paper presents \texttt{SocraticLLM} as a strong baseline to tutor students through structured conversation, encompassing review, heuristic, rectification, and summarization by integrating the Socratic method into mathematical education. By infusing extra knowledge into the LLM architecture, we ensure the reliability and quality of generated responses, thereby overcoming the issue of poor performance in complex reasoning tasks. 
To mitigate the scarcity of relevant datasets for mathematical teaching, we contribute the \texttt{SocraticMATH} dataset, comprising diverse dialogue data, enabling further advancements in this domain.
Our experiments demonstrate the efficacy of \texttt{SocraticLLM} in enhancing mathematical education. 
In further work, we would like to incorporate personal information and knowledge graphs to pave the way for future developments in adaptive and interactive learning environments.

\vspace{-1mm}
\begin{acks}
The authors wish to thank the reviewers for
their helpful comments and suggestions. 
This research is funded by the National Science and Technology Major Project (No. 2021ZD0114002), and the Science and Technology Commission of Shanghai Municipality Grant (No. 22511105901, No. 21511100402).
\end{acks}

\newpage
\clearpage
\bibliographystyle{ACM-Reference-Format}
\bibliography{acmart}

\end{document}